\documentclass[10pt,twocolumn,letterpaper]{article}

\usepackage{cvpr}              

\usepackage{algorithm}
\usepackage{algorithmic}
\usepackage{booktabs}
\usepackage{multirow} 
\usepackage{makecell}
\usepackage{subcaption}
\usepackage[table]{xcolor} 
\definecolor{Gray}{gray}{0.93}
\usepackage{float}
\definecolor{deepgreen}{RGB}{0,125,0} 
\definecolor{cvprblue}{rgb}{0.21,0.49,0.74}
\usepackage[pagebackref,breaklinks,colorlinks,allcolors=cvprblue]{hyperref}

\def\paperID{1162} 
\def\confName{CVPR}
\def\confYear{2026}

\title{Towards Generalizable AI-Generated Image Detection via \\ Image-Adaptive Prompt Learning}

\makeatletter
\newcommand{\myfnsymbol}[1]{%
  \expandafter\@myfnsymbol\csname c@#1\endcsname
}
\newcommand{\@myfnsymbol}[1]{%
  \ifcase #1
  \or \TextOrMath{\textasteriskcentered}{*}
  \or \TextOrMath{\textdagger}{\dagger}
  \fi
}
\newcommand{\equalcontributor}{\@myfnsymbol{1}}
\newcommand{\corresponding}{\@myfnsymbol{2}}
\makeatother

\author{
Yiheng Li$^{1,2}$,
Zichang Tan$^{4}$,
Guoqing Xu$^{1,2}$,
Zhen Lei$^{1,2,3,5}$,
Xu Zhou$^{4}$,
Yang Yang$^{1,2,\corresponding}$\\
$^1$ School of Artificial Intelligence, University of Chinese Academy of Sciences\\
$^2$ MAIS, Institute of Automation, Chinese Academy of Sciences\\
$^3$ CAIR, HKISI, Chinese Academy of Sciences, $^4$ Sangfor Technologies Inc.
$^5$ SCSE, FIE, M.U.S.T\\
{\tt\small \{liyiheng2024, yangyang2013, zhen.lei\}@ia.ac.cn, tanzichang@foxmail.com} 
}

\begin{document}
\maketitle
\renewcommand{\thefootnote}{\myfnsymbol{footnote}}
\footnotetext[2]{Corresponding authors}%
\begin{abstract}
In AI-generated image detection, current cutting-edge methods typically adapt pre-trained foundation models through partial-parameter fine-tuning. However, these approaches often struggle to generalize to forgeries from unseen generators, as the fine-tuned models capture only limited patterns from training data and fail to reflect the evolving traits of new ones. To overcome this limitation, we propose \textbf{I}mage-\textbf{A}daptive \textbf{P}rompt \textbf{L}earning (\textbf{IAPL}), a novel paradigm that dynamically adjusts the prompts fed into the encoder according to each testing image, rather than fixing them after training. This design significantly enhances robustness and adaptability to diverse forged images. The dynamic prompts integrate conditional information with test-time adaptive tokens through a lightweight learnable scaling factor. The conditional information is produced by a Conditional Information Learner, which leverages CNN-based feature extractors to model both forgery-specific and general conditions. The test-time adaptive tokens are optimized during inference on a single sample by enforcing prediction consistency across multiple views, ensuring that the parameters align with the current image. For the final decision, the optimal input with the highest prediction confidence is selected. Extensive experiments show that IAPL achieves state-of-the-art performance, with mean accuracies of 95.61\% and 96.7\% on the widely used UniversalFakeDetect and GenImage datasets, respectively. 
\href{https://github.com/liyih/IAPL}{Codes.}
\end{abstract}    
\section{Introduction}
\label{sec:intro}

Driven by the rapid progress in generative AI, particularly GAN \cite{goodfellow2014generative} and diffusion models \cite{song2020score}, generating high-quality synthetic images has become remarkably effortless. These breakthroughs have opened vast opportunities and applications in digital content creation, virtual reality, and personalized media. However, the potential misuse of such technologies raises serious security risks, threatening both public trust and individual privacy \cite{li2025unleashing}. This duality highlights the urgent need for robust and generalizable methods to detect AI-generated images.

\begin{figure}[t]
    \centering
    \includegraphics[width=0.46\textwidth]{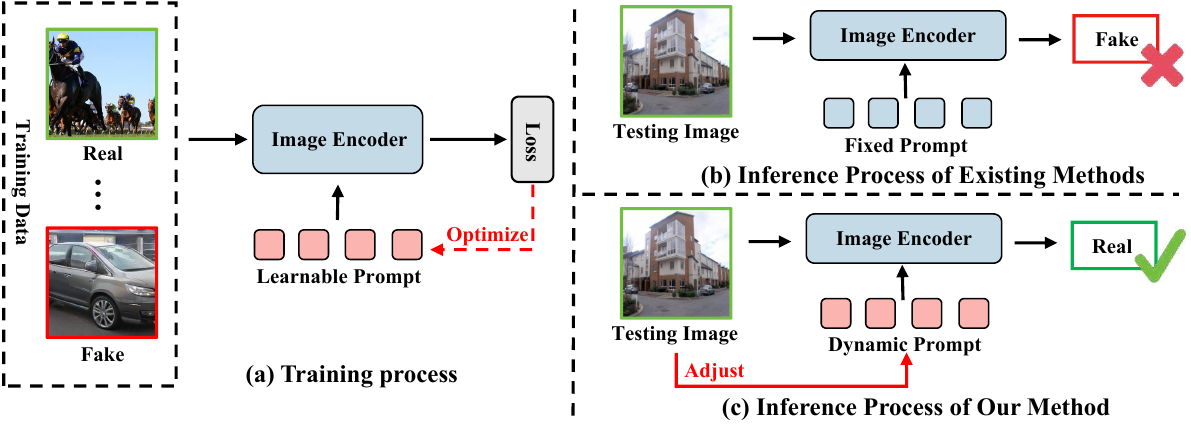}
    \caption{
    \textbf{Comparison of the adaptability of prompts between our and existing methods when facing a testing image.}
    (a) In the training process, the parameters of learnable prompts are optimized via a training set that includes limited generative methods.
    During inference, existing methods (b) keep the learned prompts fixed.
    In contrast, our method (c) dynamically adjusts the learned prompts based on test images from unseen sources, thereby enhancing generalization to novel test images.
    } 
    \label{pic0}
\end{figure} 

Several works have been proposed to detect and distinguish between AI-generated images and authentic ones. Some methods focus on detecting synthesized faces \cite{lin2025standing, yan2024transcending}, while others focus on detecting synthesized natural images \cite{zhu2023genImage, yan2024sanity}. Beyond recognizing images generated by seen generators in the training set, a more important task is detecting images produced by unseen generators. The significance lies in improving the model's generalizability to the ever-evolving generation methods emerging in real-world scenarios. To this end, certain approaches find cues from the low-level traces in the images, such as frequency \cite{tan2024frequency}, neighborhood pixel \cite{tan2024rethinking}, and gradients \cite{tan2023learning}. However, these methods rely on local details, to some extent overlooking the contribution of unrealistic semantic patterns and overall style in AI-generated content.

Recent state-of-the-art (SOTA) approaches tend to fine-tune visual foundation models, such as CLIP \cite{radford2021learning}, to boost detection performance. 
These pre-trained models encode rich real-world knowledge, which serves as a valuable complement to the limited fake patterns present in training data.
Nevertheless, fine-tuned models often struggle to generalize effectively under domain shifts introduced by images from unseen generators, revealing limitations in current adaptation strategies.
We contend that achieving stronger generalization requires learnable parameters that remain adaptable to diverse image characteristics.
Since images synthesized by different models exhibit substantial variations in texture, semantics, and visual artifacts, relying on fixed model parameters is inadequate for capturing the essential discriminative properties.
In this view, we propose a novel \textbf{I}mage-\textbf{A}daptive \textbf{P}rompt \textbf{L}earning (\textbf{IAPL}) method. As shown in Fig. \ref{pic0}, the existing methods \cite{shi2025mirage, lin2025standing} keep the prompts learned during training fixed when inferring testing images from unseen sources. In contrast, for IAPL, the prompts input to the image encoder are dynamically adjusted based on the traits of the image during inference, which enables the model to capture instance-specific cues and better handle variations introduced by different generative methods.

\begin{figure}
    \centering
    \includegraphics[width=0.45\textwidth]{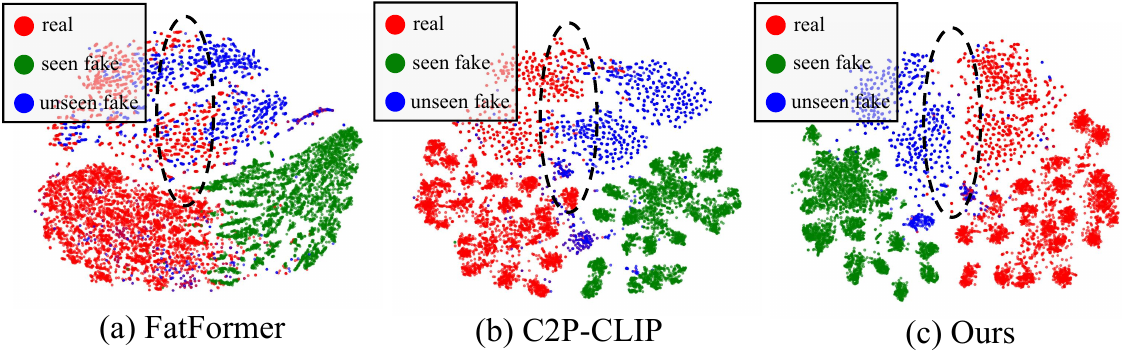}
    \caption{
    \textbf{T-SNE visualization between \textbf{\textcolor{red}{real}}, \textbf{\textcolor{deepgreen}{seen fake}} and \textbf{\textcolor{blue}{unseen fake}} images on UniversalFakeDetect dataset.} Compared to C2P-CLIP and Fatformer, our approach yields features of \textbf{\textcolor{blue}{unseen fake}} images that are more similar to those of \textbf{\textcolor{deepgreen}{seen fake}} images and more distinct from \textbf{\textcolor{red}{real}} images. As shown by the \textbf{black dashed area}, our method produces a more distinct boundary between the features of \textbf{\textcolor{blue}{unseen fake}} and \textbf{\textcolor{red}{real}} images. For FatFormer, we use the augmented image features from Text-Guided Interactor.
    } 
    \label{pic1}
\end{figure} 

The image-adaptive prompts are input into the first encoder block and consist of test-time adaptive tokens and conditional information. During training, test-time adaptive tokens are optimized in the same way as standard learnable tokens. At inference, however, they are adjusted based on a single test image, minimizing entropy across different generated views to alleviate the uncertainty introduced by domain shift \cite{shu2022test}. The conditional information is processed by the Conditional Information Learner, which leverages high-pass filters and CNN-based feature extractors to capture both forgery-specific and general cues from local regions rich in texture. These conditions are instance-dependent, varying across images, and are injected into the detection pipeline to guide feature learning and capture subtle artifacts. 
To better fuse test-time adaptive token and conditional information, we further introduce the learnable scaling factor, a control module that adaptively regulates information flow. It aggregates the two types of information using learnable coefficients, emphasizing relevant signals and suppressing less informative ones.
Moreover, to equip the pre-trained foundation model with enhanced task-specific capabilities for AI-generated image detection, we also insert some learnable parameters for fine-tuning, which will be fixed after training. Here, we combine MLP-based adapters \cite{chen2022adaptformer} and learnable tokens \cite{zhou2022learning}, enabling efficient adaptation while preserving the generalization capability of the backbone model. The former inserts learnable upsampling and downsampling matrices alongside the MLP module in the encoder blocks, which are placed at equal intervals throughout the encoder. The latter is applied to the encoder blocks from the 2$^{nd}$ to the $N_t$-th. 

We conduct extensive experiments on two widely used datasets, UniversalFakeDetect \cite{ojha2023towards} and GenImage \cite{zhu2023genImage}, proving the effectiveness of our method. Specifically, it achieves 95.61\% and 96.7\% mAcc on these two datasets, respectively, reaching the new SOTA performance. Ablation studies confirm that all proposed components play a role in enhancing performance. As shown in Fig. \ref{pic1},  our method achieves more distinct separation between real and unseen fake images, demonstrating its strong generalization ability. Contributions are as follows:
\begin{itemize}
\item We propose a novel Image-Adaptive Prompt Learning strategy, aiming to enhance the model's generalizability to wider images generated by unseen generators.
\item We develop an efficient adaptation scheme for pre-trained vision-language models, combining lightweight MLP-based adapters with learnable tokens. This design maintains the backbone’s ability to extract features while allowing flexible adaptation.
\item Extensive experiments demonstrate that our method exhibits strong generalizability across different generators.
\end{itemize}
\section{Related Work}
\label{sec:related}

\subsection{Face Forgery Detection}
To ensure the privacy and security of users, face forgery detection has become an important task. Most existing works design sophisticated transformation functions to highlight the fake patterns in generated faces. These transformations function includes frequency-domain transformation \cite{li2021frequency, frank2020leveraging}, spatial-domain operation \cite{wang2023noise}, reconstruction technique \cite{cao2022end} and disentanglement \cite{fu2025exploring}. Other works focus on capturing artifacts from facial regions with prominent textures, such as the lips \cite{haliassos2021lips} and eyes \cite{li2018ictu}. For example, LAA-Net \cite{nguyen2024laa} targets small, artifact-prone regions and propagates discriminative low-level features into the final output.
Nevertheless, as the variety of generative models continues to expand, the effectiveness of methods in handling forgeries from unseen sources remains limited. Thus, some works adopt techniques like blending \cite{shiohara2022detecting} and adversarial training \cite{chen2022self} to enrich the training samples. For example, by modeling and simulating intra-forgery and inter-forgery feature variations in the latent space, LSDA \cite{yan2024transcending} enlarges the forgery space and mitigates the risk of the model overfitting to forgery-specific artifacts.

\subsection{AI-Generated Image Detection}
In recent years, generative content has expanded beyond human faces to areas like nature and art.
Several approaches capture clues from low-level artifacts, including frequency-based features \cite{jeong2022bihpf}, reconstruction artifacts \cite{wang2023dire}, spatial pixel relationships \cite{tan2024rethinking}, and gradients \cite{tan2023learning}. For example, BiHPF \cite{jeong2022bihpf} leverages two high-pass filters in both the frequency and pixel domains to extract forgery cues. Another important paradigm \cite{liu2026mirror,liu2025beyond,qin2025Scaling} is to leverage the vision foundation models. These methods benefit from the extensive pre-trained knowledge in visual foundation models and extract high-level forgery features. UniFD \cite{ojha2023towards} is the pioneer to use the CLIP \cite{radford2021learning} features for forgery detection. FatFormer \cite{liu2024forgery} conducts forgery-aware adapter with frequency analysis to fine-tune CLIP.
RINE \cite{koutlis2024leveraging} maps the intermediate features of CLIP to a learnable forgery-aware space through a lightweight network.
C2P-CLIP \cite{tan2025c2p} injects category-related concepts to the encoder via contrastive learning.
Effort \cite{yan2024orthogonal} introduces orthogonal subspace decomposition to separate domain-specific and domain-invariant features.
Additionally, some works \cite{yan2024sanity, cheng2025co} combine low-level and high-level features for more robust representation.

\begin{figure}[ht]
  \centering
   \includegraphics[width=0.46\textwidth]{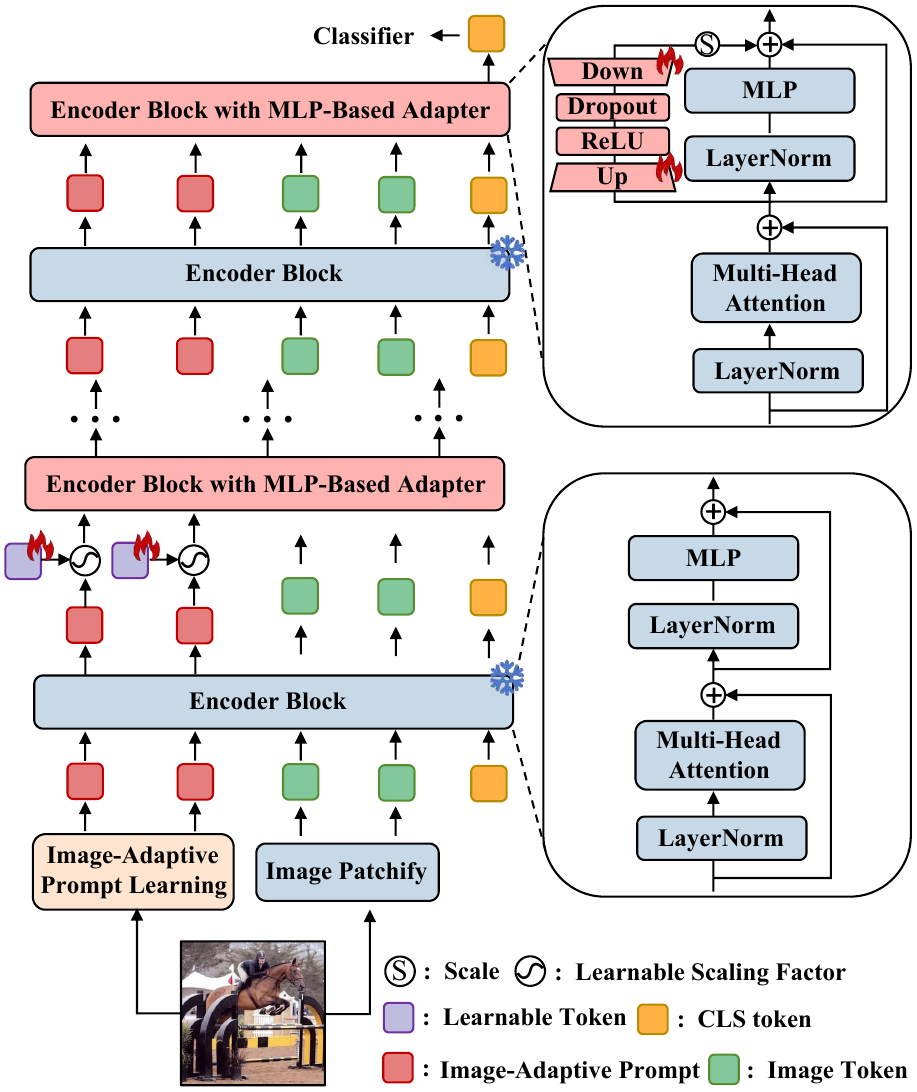}
   \caption{\textbf{Overview of the detection pipeline in our method.} The trainable parameters added to the original CLIP ViT in our image encoder fall into three categories: MLP-based adapters, learnable tokens, and image-adaptive prompts. MLP-based adapters go into $N_a$ encoder blocks at regular intervals. Learnable tokens are used from the $2^{nd}$ to the $N_t$-th encoder blocks. Image-adaptive prompts, learned via their dedicated module, are applied as the first block's input. After training, MLP-based adapters and learnable tokens are fixed, while image-adaptive prompts are adjusted based on the image during inference.
   }
   \label{fig2}
\end{figure}

\subsection{Prompt Learning}

Prompt learning is initially introduced to substitute handcrafted text templates with learnable tokens \cite{liu2021p, zhou2022learning}. Subsequently, an increasing number of studies began to explore its effectiveness in the visual modality. VPT \cite{jia2022visual} inserts a small number of learnable tokens into the pretrained ViT \cite{dosovitskiy2020image} layers to adapt the model to downstream tasks. VP \cite{bahng2022exploring} uses a learnable image perturbation to be the visual prompt. Then, some multi-modal prompt learning methods are introduced for a visual-language model \cite{khattak2023maple}. Moreover, to improve generalization to broader unseen classes, methods like input-conditional learning and test-time adaptation \cite{wang2020tent} have been proposed. CoCoOp \cite{zhou2022conditional} adds input-conditional tokens to each instance and makes the model less sensitive to class shift. TPT \cite{shu2022test} is the pioneer to do Test-Time Tuning in prompt learning by minimizing the averaged entropy to enforce consistent predictions across selected augmentations. R-TPT \cite{sheng2025r} chooses to minimize pointwise entropy to reduce conflicts.

Several prior studies have explored the application of prompt learning \cite{liu2024forgery, lin2025standing} in the context of synthetic image detection. However, the prompts learned by these methods do not generalize well to images generated by unseen generators. In this view, we propose an Image-Adaptive Prompt Learning strategy, which dynamically adjusts the learned prompt based on the characteristics of the test image and increases the adaptability to domain shift.   

\section{Method}

\subsection{Overview}

\begin{figure*}[t]
  \centering
   \includegraphics[width=0.94\textwidth]{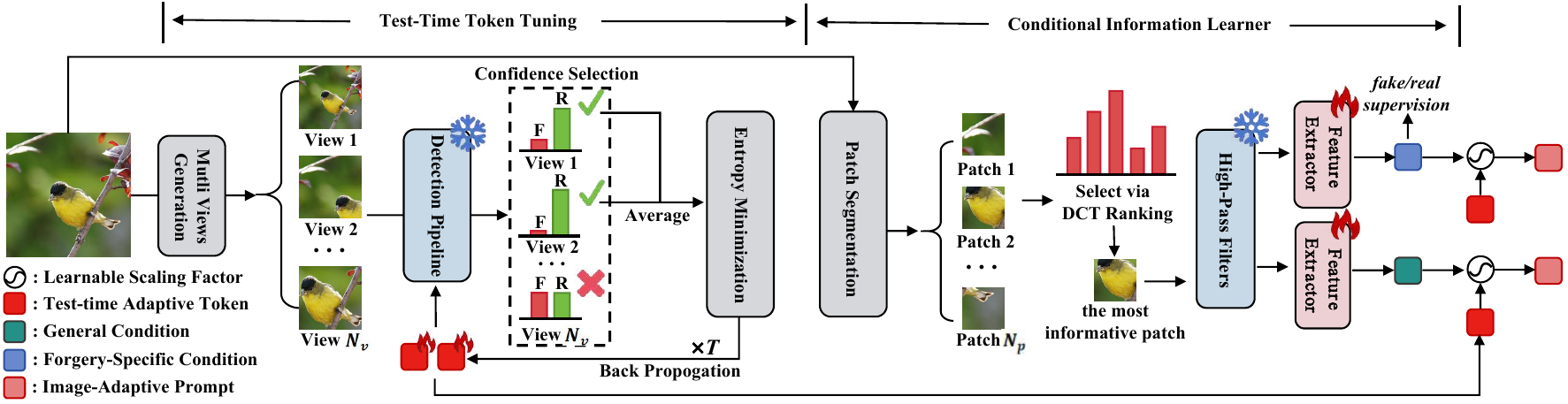}
   \caption{\textbf{The overall architecture of Image-Adaptive Prompt Learning.} It consists of Test-Time Token Tuning and Conditional Information Learner. The former dynamically adjusts test-time adaptive tokens based on the input image, while the latter extracts conditional information from the texture-rich region, providing forgery-specific and general guidance for prompt generation.}
   \label{fig3}
\end{figure*}

The overview of our detection pipeline is shown in Fig. \ref{fig2}. We design a flexible yet effective framework that tailors the pre-trained CLIP \cite{radford2021learning} ViT for AI-generated image detection by integrating three trainable components: MLP-based adapters, learnable tokens, and image-adaptive prompts. These components preserve the backbone’s discriminative power while enabling instance-specific adaptation to improve robustness against unseen generators.

Given an input image, it is first fed into the Image-Adaptive Prompt Learning module to obtain image-adaptive prompts that encode image-oriented cues to guide the backbone model. What's more, following the vanilla CLIP, the input image is divided into multiple patches and encoded into image tokens $f_{img} \in \mathbb{R}^{N \times D}$, where $N$ and $D$ are the number of tokens and dimension, respectively. After that, the generated image-adaptive prompts, image tokens, and additional CLS token $f_{cls} \in \mathbb{R}^{1 \times D}$, are fed into multiple encoder blocks for feature extraction. In each block, the pre-trained parameters of the original CLIP are fixed to preserve knowledge. MLP-based adapters are inserted at equal intervals into $N_a$ blocks, while learnable tokens are applied from the $2^{nd}$ to the $N_t$-th block. 
In blocks with learnable tokens, the prompts from the previous block are fused with the tokens via a learnable scaling factor to form the current input prompts, thereby propagating adaptive information from the image-adaptive prompts and strengthening forgery cue modeling in deeper layers. Finally, the output CLS token from the last encoder block is used for the result prediction via a trainable classifier.

\subsection{Fixed Learned Parameters}

We first introduce the MLP-based adapters and learnable tokens, which remain unchanged after training and do not vary with input images. Therefore, they are referred to as fixed learned parameters. It is worth noting that, although our paper emphasizes the importance of dynamic parameters in capturing instance-specific forgery cues, the fixed learned parameters still play a crucial role: they provide a stable backbone for the model, allowing dynamic parameters to focus on adaptive adjustments for each input image while preventing excessive flexibility that could compromise overall inference accuracy. 

\textbf{MLP-based Adapter.} 
As shown in Fig. \ref{fig2}, we employ a simple yet effective MLP-based adapter \cite{chen2022adaptformer}. It includes a down-projection $W_{d} \in \mathbb{R}^{D \times \hat{D}}$ and a up-projection $W_{d} \in \mathbb{R}^{\hat{D} \times D}$, where $\hat{D}$ is the middle dimension and satisfies $\hat{D} \ll D$. A ReLU activation function and dropout technique are employed between the up-projection and down-projection to enhance non-linearity and prevent overfitting. The adapter’s output is scaled by a factor $s$ and subsequently added to the original output. For parameter initialization, we follow LoRA \cite{hu2022lora}: the down-projection is initialized using Kaiming uniform initialization \cite{he2015delving}, while the up-projection is initialized to zeros.

\textbf{Learnable Token.} Formally, in the $j$-th block $(2 \leq j \leq N_t)$, a set of learnable tokens 
$T^{(j)} \in \mathbb{R}^{M \times D}$ is fused with the output prompts 
$P_o^{(j-1)} \in \mathbb{R}^{M \times D}$ from the former block through a learnable scaling factor to produce the 
input prompts $P_i^{(j)} \in \mathbb{R}^{M \times D}$ of this block. Here, $M$ denotes the number of learnable tokens, and $D$ is the embedding dimension. The learnable scaling factor is defined in Eq. \ref{eq1}.
\begin{equation}
\label{eq1}
P_i^{(j)} = \alpha^{(j)} * {P_o}^{(j-1)} + T^{(j)},
\end{equation}
where $\alpha^{(j)} \in \mathbb{R}^{D}$ denotes the channel-wise coefficients in the $j$-th layer. These coefficients are initialized to small values and optimized during training. They are applied directly to the prompt, enabling fine-grained control over how information from the previous layer is integrated on a per-channel basis. In other words, $\alpha^{(j)}$ modulates the contribution of each channel in the fused prompt, allowing the model to emphasize or suppress specific features dynamically. Then, the generated prompts $P_i^{(j)}$ are concatenated with the image and CLS tokens and fed into the subsequent encoder blocks for further representation learning.

\subsection{Image-Adaptive Prompt Learning}
While previously mentioned adapters and learnable tokens provide a stable backbone for detection, they lack the flexibility to adapt to diverse image characteristics, especially when encountering content generated by unseen models. Given that AI-generated images often vary in texture, semantics, and artifacts, relying solely on static parameters makes the model vulnerable to domain shifts. In this view, as shown in Fig. \ref{fig3}, we introduce Image-Adaptive Prompt Learning, which dynamically adjusts input prompts according to the characteristics of each test image. 

\textbf{Test-Time Token Tuning.} The dynamic adaptability of this module is primarily embodied in the test-time adaptive tokens. Each token is a trainable embedding that is optimized in the same manner as standard learnable tokens during training. However, at inference time, its parameters are further adjusted according to a single testing image, enabling instance-specific adaptation and enhancing robustness to domain shifts.

Specifically, at the inference stage, the Test-Time Token Tuning starts with generating multiple views $V \in R^{N_v\times224\times224\times3}$ from the original image, where $N_v$ is the number of views. In particular, the generated views could be divided into a global view and multiple local views. For a global view, we resize the entire image to the input size of CLIP, i.e. $224\times224$. For local views, we randomly crop $N_v$-1 $224\times224$ sub-images followed by horizontal flipping augmentation. If the image is too small to crop $N_v$-1 different sub-images, we randomly crop $N_v$-1 smaller square patches and then resize these patches to $224\times224$. To reduce the noise from randomly generated views, the Confidence Selection is conducted. We select $m$ views with high predicted confidence for subsequent tuning. The predicted confidence $S_c^i$ is calculated via Eq. \ref{eq2}.
\begin{equation}
\label{eq2}
S_c^i = 2 *|\sigma(z_i)-0.5|,
\end{equation}
where $z_i$ is the logit of the i-th view. $\sigma(.)$ is the Sigmoid function that transform $z_i$ into $(0, 1)$.

Given the selected views, we tune the test-time adaptive tokens by minimizing entropy, while other parameters are frozen. The objective function mainly includes  
averaged entropy loss \cite{shu2022test} and pointwise entropy loss \cite{sheng2025r}.
For the averaged entropy loss, its purpose is to improve consistency among predictions across the generated views, and it is shown in Eq. \ref{eq3} and Eq. \ref{eq4}.
\begin{equation}
\label{eq3}
L_{avg} = -(\overline{p}*log(\overline{p})+(1-\overline{p})*log(1-\overline{p})),
\end{equation}
\begin{equation}
\label{eq4}
\overline{p}=\frac{1}{m} \sum_{k=1}^{m} \sigma(z_k),
\end{equation}
Here, $z_k$ is the logit calculated by the model of the $k$-th selected view.
Some methods \cite{sheng2025r} adopt pointwise entropy loss to push the decision boundary away from low-entropy regions. Its objective function is shown in Eq. \ref{eq5}.
\begin{equation}
\label{eq5}
L_{point}= \frac{1}{m}\sum_{k=1}^{m}-(p_k*log(p_k)+(1-p_k)*log(1-p_k)),
\end{equation}
where $p_k$ denotes $\sigma(z_k)$ of k-th selected view. 
Based on the results in Tab. \ref{tab:8}, we ultimately adopt the averaged entropy loss due to its superior performance across all experimental datasets.
Additionally, the above tuning process is performed for a total of $T$ steps.

\textbf{Conditional Information Learner.} 
To capture subtle and varying cues in AI-generated images, we propose the Conditional Information Learner, which extracts conditional information tailored to each input. The conditional information is partitioned into forgery-specific and general components, which act on the input prompt to enable the model to focus on the distinct characteristics of various inputs.
Moreover, the condition is obtained from the most texture-rich region with a high-pass filter, emphasizing low-level patterns often overlooked by the high-level semantic features of the pre-trained CLIP backbone. 

Specifically, for a given input image, we divide it into local patches $P_l \in R^{N_p \times 32 \times 32 \times 3}$, where $N_p$ is the number of patches. We use the DCT score \cite{yan2024sanity} to measure the texture richness of each patch and select the patch with the highest DCT value for conditional information extraction. After that, the selected patches are input into high-pass filters \cite{fridrich2012rich} to obtain the high-frequency pattern. 
Two CNN-based feature extractors with identical architectures but separate parameters are used to extract the forgery-specific condition $C_f$ and the general condition $C_g$, respectively.
An auxiliary supervision branch is attached after a forgery-specific condition. It consists of a single-layer MLP, enabling the supervised branch to extract features strongly correlated with the presence of forgery, while the unsupervised branch learns more general features related to the image status. Moreover, as the conditional information is extracted from a small patch, this process is both time-efficient and memory-efficient.

After acquiring the conditional information $C_g$ and $C_f$, we fuse them with the test-time adaptive tokens $A \in \mathbb{R}^{2 \times D}$ via the learnable scaling factor, to obtain the image-adaptive prompt $P \in \mathbb{R}^{2 \times D}$ as shown in Eq. \ref{eq6}.
\begin{equation}
\label{eq6}
P = \{\alpha_f * C_f + A [0,:], \alpha_g * C_g + A [1,:]\},
\end{equation}
where $\alpha_f$ and $\alpha_g$ denote the learnable coefficients associated with the forgery-specific and general conditional information, respectively, and $\{\,\cdot\,,\,\cdot\,\}$ represents the concatenation operation.

\textbf{Loss Function.} In the train phase, the overall loss $L_{overall} = L_{cls} + L_{aux}$. $L_{cls}$ is the final classification loss in the classifier, while $L_{aux}$ is the auxiliary loss in the Conditional Information Learner. Both of them are computed based on the binary cross-entropy loss. In the test phase, the averaged entropy loss $L_{avg}$ is used to tune the test-time adaptive tokens.

\subsection{Optimal Input Selection}

Since images generated by different models often vary considerably in resolution and aspect ratio, previous methods typically adopt resizing or center cropping to standardize the input. However, such fixed preprocessing strategies may discard or distort critical artifacts, as the most discriminative regions for forgery detection differ across images. 
To address this limitation, we propose Optimal Input Selection, which only occurs during inference. 
Specifically, we first generate diverse inputs for a single test image. After testing-time token tuning, these inputs are fed into the tuned model to yield predictions.
Their confidence scores are computed using the same procedure as Eq.~\ref{eq2}, and the prediction with the highest score is taken as the final decision. Additionally, we reuse the $m$ views selected through confidence selection as the inputs for convenience. 

\section{Experiments} 
\subsection{Datasets and Metrics}
We conduct experiments on two widely used datasets, i.e., UniversalFakeDetect Dataset \cite{ojha2023towards} and GenImage Dataset \cite{zhu2023genImage}, to demonstrate the effectiveness of our framework.

\textbf{UniversalFakeDetect Dataset.} This dataset includes 20 subsets of generated images. Following previous protocol \cite{ojha2023towards, wang2020cnn}, the real and fake images in ProGAN \cite{progan} are used for training, while the testing set contains 19 subsets from ProGAN and other 18 different generative models, including CycleGAN \cite{Zhu-2017-cycleGAN}, BigGAN \cite{brock2018brock}, StyleGAN \cite{stylegan}, GauGAN \cite{park2019SPADE}, StarGAN \cite{choi2018stargan}, DeepFakes~\cite{roessler2019faceforensicspp}, SITD \cite{sitd}, SAN \cite{san}, CRN \cite{crn}, IMLE \cite{imle}, Guided \cite{dhariwal2021diffusion}, LDM \cite{rombach2022high}, Glide \cite{glide}, and DALLE \cite{dalle-orig}.

\textbf{GenImage Dataset.} This dataset mainly explores the Diffusion model for AI-generated image detection. Following its original setting \cite{zhu2023genImage}, we adopt SD v1.4 as the training set and the remaining sets as the testing set, 
including Midjourney \cite{Midjourney}, SDv1.4 \cite{rombach2022high}, SDv1.5 \cite{rombach2022high}, ADM \cite{dhariwal2021diffusion}, GLIDE \cite{glide}, Wukong \cite{Wukong}, VQDM \cite{gu2022vector}, and BigGAN \cite{brock2018brock}. As the varied image size of GenImage, following \cite{tan2025c2p}, we duplicate the images whose size is smaller than 224 and crop them to $224\times224$ during training.

\textbf{Metrics.} Following the convention in C2P-CLIP\cite{tan2025c2p}, we use average precision (AP) and classification accuracy (Acc) for evaluation on the UniversalFakeDetect dataset, and only Acc on the GenImage dataset, with a 0.5 threshold for Acc across all datasets.

\subsection{Implementation Details} 

Following \cite{ojha2023towards}, we adopt CLIP ViT-L/14 as the image encoder. During training, we choose Adam \cite{kingma2014adam} as the optimizer with the learning rate of $5\times10^{-5}$. The batch size is set to 16 on UniversalFakeDetect and 32 on GenImage. The model is trained on a single NVIDIA 3090 GPU with only one epoch. The feature dimension $D$ of the image encoder is set to 1024. $N_a$ is set to 6 and $N_t$ is set to 9.
For the MLP-based adapter, the middle dimension $\hat{D}$ is set to 64, and the scale factor $s$ is set to 0.1. For the Conditional Information Learner, the patch number $N_p$ is set to 192. The number of learnable tokens $M$ inserted in each block is set to 2. For the Test-Time Token Tuning, the number of views $N_v$ is set to 32, the number of selected confident views $m$ is set to 6, the tuning step number $T$ is set to 2, and the learning rate during Test-Time Token Tuning is set to $5\times10^{-3}$.

\begin{table*}[!ht]
\caption{\textbf{Evaluation on the UniversalFakeDetect dataset \cite{ojha2023towards} using Accuracy (\%).} The results of RINE are cited from their original report. The results of AIDE are reproduced by its official pre-trained model. Results of the other methods are cited from C2P-CLIP. $\clubsuit$: evaluating based on the official pre-trained model. \textbf{The 4-class ProGAN is used as the training data}.
}
    \centering
    \tabcolsep=0.1cm
\resizebox{\textwidth}{!}{
    \begin{tabular}{c|c|c|c|c|c|c|c|c|c|c|c|c|c|c|c|c|c|c|c|c}
    \bottomrule 
      \multirow{3}*{Methods}  & \multicolumn{6}{c|}{GAN} &  \multirow{3}*{\makecell[c]{Deep\\fakes}} & \multicolumn{2}{c|}{Low level} & \multicolumn{2}{c|}{Perceptual loss} &  \multirow{3}*{Guided} & \multicolumn{3}{c|}{LDM} & \multicolumn{3}{c|}{Glide} &  \multirow{3}*{Dalle}  &  \multirow{3}*{mAcc}\\ 
       \cline{2-7} \cline{9-10}  \cline{11-12}  \cline{14-16}  \cline{17-19}  
       ~ & \makecell[c]{Pro-\\GAN} & \makecell[c]{Cycle-\\GAN} & \makecell[c]{Big-\\GAN} & \makecell[c]{Style-\\GAN}  & \makecell[c]{Gau-\\GAN}  & \makecell[c]{Star-\\GAN} & ~ & \makecell[c]{SITD}& \makecell[c]{SAN}& \makecell[c]{CRN}& \makecell[c]{IMLE}& ~ & {\makecell[c]{200\\steps}}& {\makecell[c]{200\\w/cfg}}& {\makecell[c]{100\\steps}}& {\makecell[c]{100\\27}} & {\makecell[c]{50\\27}} & \makecell[c]{100\\10} & ~ & ~\\
       \hline
       UniFD~\cite{ojha2023towards} & 100.00 & 98.50 & 94.50 & 82.00 & \underline{99.50} & 97.00 & 66.60 & 63.00 & 57.50 & 59.50 & 72.00 & 70.03 & 94.19 & 73.76 & 94.36 & 79.07 & 79.85 & 78.14 & 86.78 & 81.38 \\
       LGrad~\cite{tan2023learning} & 99.84 & 85.39 & 82.88 & 94.83 & 72.45 & 99.62 & 58.00 & 62.50 & 50.00 & 50.74 & 50.78 & 77.50 & 94.20 & 95.85 & 94.80 & 87.40 & 90.70 & 89.55 & 88.35 & 80.28  \\
       FreqNet~\cite{tan2024frequency} & 97.90 & 95.84 & 90.45 & 97.55 & 90.24 & 93.41 & \bf97.40 & 88.92 & 59.04 & 71.92 & 67.35 & \underline{86.70} & 84.55 & \bf99.58 & 65.56 & 85.69 & 97.40 & 88.15 & 59.06 & 85.09 \\ 
       NPR~\cite{tan2024rethinking} & 99.84 & 95.00 & 87.55 & 96.23 & 86.57 & 99.75 & 76.89 & 66.94 & \bf98.63 & 50.00 & 50.00 & 84.55 & 97.65 & 98.00 & 98.20 & 96.25 & 97.15 & 97.35 & 87.15 & 87.56 \\
       FatFormer~\cite{liu2024forgery} & 99.89 & \bf{99.32} & \underline{99.50} & 97.15 & 99.41 & 99.75 & 93.23 & 81.11 & 68.04 & 69.45 & 69.45 & 76.00 & 98.60 & 94.90 & 98.65 & 94.35 & 94.65 & 94.20 & 98.75 & 90.86 \\
       RINE~\cite{koutlis2024leveraging}&\underline{100.00}& \underline{99.30}& \bf99.60& 88.90& \bf99.80& 99.50 &80.60 &90.60 &68.30 &89.20 &90.60& 76.10& 98.30& 88.20& 98.60& 88.90& 92.60& 90.70& 95.00 &91.31\\
       AIDE$^\clubsuit$ ~\cite{yan2024sanity} &99.99&98.49&83.95 &\bf99.65&73.24&\underline{99.90}&54.09&68.06&71.14&60.87&61.00&\bf88.50&98.20&97.45&98.40&\bf98.15&\bf98.40&97.85&97.50& 86.57 \\
       C2P-CLIP ~\cite{tan2025c2p} &99.98& 97.31& 99.12& 96.44& 99.17& 99.60& 93.77& \bf95.56& 64.38& \bf93.29& \bf93.29& 69.10& \underline{99.25}& 97.25& \bf99.30& 95.25& 95.25& 96.10& 98.55& \underline{93.79} \\
       
       SAFE$^\clubsuit$~\cite{li2025improving}&99.86&98.87&89.73&\underline{98.04}&91.52&\bf99.90&93.10&85.56&\underline{95.91}&50.10&50.10&82.35&98.80&\underline{98.65}	&98.80&95.75&96.60&97.25&97.50&90.44\\
       MiraGe~\cite{shi2025mirage}&100.00&94.30&96.50&96.80&93.60&96.10&88.70&75.80&71.90&\underline{92.90}&\underline{92.90}&82.00&98.30&94.60&98.60&97.50&97.50&\underline{98.00}&\underline{98.60}&92.90 \\
       \hline
       \rowcolor{blue!20} Ours &\bf100.00&98.60&98.65&94.89&99.39&96.70&\underline{95.89}&\underline{90.83}&93.84 &92.47&92.72&72.75&\bf99.50&97.70&\underline{99.15}&\underline{97.95}&\underline{98.30}&\bf98.35&\bf98.90&\textbf{95.61} \\
\bottomrule
    \end{tabular}
}
  \label{tab:1}
\end{table*}

\begin{table*}[!ht]
\caption{\textbf{Evaluation on the UniversalFakeDetect dataset \cite{ojha2023towards} using Average Precision (\%).}  The results of previous methods are cited from the same sources as Table \ref{tab:1}. \textbf{The 4-class ProGAN is used as the training data}.}
    \centering
    \tabcolsep=0.1cm
\resizebox{\textwidth}{!}{
    \begin{tabular}{c|c|c|c|c|c|c|c|c|c|c|c|c|c|c|c|c|c|c|c|c}
    \bottomrule 
      \multirow{3}*{Methods}  & \multicolumn{6}{c|}{GAN} &  \multirow{3}*{\makecell[c]{Deep\\fakes}} & \multicolumn{2}{c|}{Low level} & \multicolumn{2}{c|}{Perceptual loss} &  \multirow{3}*{Guided} & \multicolumn{3}{c|}{LDM} & \multicolumn{3}{c|}{Glide} &  \multirow{3}*{Dalle}  &  \multirow{3}*{mAP}\\ 
      
       \cline{2-7} \cline{9-10}  \cline{11-12}  \cline{14-16}  \cline{17-19}  
       ~ & \makecell[c]{Pro-\\GAN} & \makecell[c]{Cycle-\\GAN} & \makecell[c]{Big-\\GAN} & \makecell[c]{Style-\\GAN}  & \makecell[c]{Gau-\\GAN}  & \makecell[c]{Star-\\GAN} & ~ & \makecell[c]{SITD}& \makecell[c]{SAN}& \makecell[c]{CRN}& \makecell[c]{IMLE}& ~ & {\makecell[c]{200\\steps}}& {\makecell[c]{200\\w/cfg}}& {\makecell[c]{100\\steps}}& {\makecell[c]{100\\27}} & {\makecell[c]{50\\27}} & \makecell[c]{100\\10} & ~ & ~\\
       \hline
       UniFD~\cite{ojha2023towards} & 100.00 & 98.13 & 94.46 & 86.66 & 99.25 & 99.53 & 91.67 & 78.54 & 67.54 & 83.12 & 91.06 & 79.24 & 95.81 & 79.77 & 95.93 & 93.93 & 95.12 & 94.59 & 88.45 & 90.14 \\
       LGrad~\cite{tan2023learning} & 100.00 & 93.98 & 90.69 & 99.86 & 79.36 & 99.98 & 67.91 & 59.42 & 51.42 & 63.52 & 69.61 & 87.06 & 99.03 & 99.16 & 99.18 & 93.23 & 95.10 & 94.93 & 97.23 & 86.35  \\
       FreqNet~\cite{tan2024frequency} & 99.92 & 99.63 & 96.05 & 99.89 & 99.71 & 98.63 & \bf99.92 & 94.42 & 74.59 & 80.10 & 75.70 & 96.27 & 96.06 & \bf100.00 & 62.34 & 99.80 & 99.78 & 96.39 & 77.78 & 91.95 \\ 
       NPR~\cite{tan2024rethinking} &  100.00 & 99.53 & 94.53 & \underline{99.94} & 88.82 & 100.00 & 84.41 & \underline{97.95} & \bf99.99 & 50.16 & 50.16 & \bf98.26 & 99.92 & 99.91 & 99.92 & \bf99.87 & \bf99.89 & \bf99.92 & 99.26 & 92.76  \\
       FatFormer~\cite{liu2024forgery} & 100.00 & 100.00 & \bf99.98 & 99.75 & 100.00 & 100.00 & 97.99 & 97.94 & 81.21 & 99.84 & 99.93 & 91.99 & 99.81 & 99.09 & 99.87 & 99.13 & 99.41 & 99.20 & 99.82 & 98.16 \\
        RINE~\cite{koutlis2024leveraging}&100.00&100.00&99.90&99.40&100.00& 100.00&  97.90& 97.20& 94.90& 97.30& 99.70& 96.40& 99.80& 98.30& 99.90& 98.80& 99.30& 98.90& 99.30& \underline{98.79}\\
        AIDE$^\clubsuit$ ~\cite{yan2024sanity} &100.00&99.89&94.44&\bf100.00&97.69&100.00 &76.35&83.60&88.82&90.73&99.94&97.28&99.29&99.12&99.34&99.30&99.25&99.19&99.03& 95.96\\
       C2P-CLIP ~\cite{tan2025c2p} &100.00& \underline{100.00}& \underline{99.96}& 99.50& \underline{100.00}& 100.00& \underline{98.59}& \bf98.92& 84.56& \underline{99.86}& 99.95& 94.13& \bf99.99& 99.83& \bf99.98& 99.72& \underline{99.79}& 99.83& \underline{99.91}& 98.66\\
       SAFE$^\clubsuit$~\cite{li2025improving}&100.00&99.77&95.88&99.93&97.19&\underline{100.00}&97.45&85.07&\underline{99.27}&41.32&47.23&95.81&\underline{99.96}&\underline{99.92}&\underline{99.96}&98.93&99.21&99.36&99.67&92.42\\
       MiraGe~\cite{shi2025mirage}&\underline{100.00}&\bf100.00&99.90&99.80&99.90&99.90&96.00&93.90&84.70&99.90&\underline{100.00}&\underline{96.40}&99.90&99.10&99.90&\underline{99.80}&99.70&99.80&99.90&98.30 \\
      \hline
       \rowcolor{blue!20} Ours  &  \bf100.00&  99.99 &99.95  &99.75  &\bf100.00  &\bf100.00  &97.59  &97.27  &98.12  &\bf99.97  &\bf100.00  &96.25  &99.86  &99.60  &99.74  &99.40  &99.73  &\underline{99.86}  &\bf99.95& \textbf{99.32} \\
\bottomrule
    \end{tabular}
}
  \label{tab:2}
\end{table*}

\begin{table*}[!ht]
\caption{\textbf{Evaluation on the GenImage \cite{zhu2023genImage} dataset using Accuracy (\%).} The results of AIDE, DRCT, and ATTSD are cited for their original reports. $\spadesuit$: re-implementing on SDv1.4 setting. The others are cited from C2P-CLIP. \textbf{The SDv1.4 is used as the training set}. }
 \centering
 \tabcolsep=0.38cm
 \resizebox{\textwidth}{!}{
    \begin{tabular}{c c c c c c c c c c | c}
    \bottomrule \hline
        Methods & Venues & Midjourney & SDv1.4 & SDv1.5 & ADM & GLIDE & Wukong & VQDM & BigGAN & mAcc\\
          \bottomrule \hline
UniFD~\cite{ojha2023towards}  & CVPR 2023  & 91.5&96.4&96.1&58.1&73.4&94.5&67.8&57.7 &79.5\\
NPR ~\cite{tan2024rethinking}  &  CVPR 2024 & 81.0 & 98.2 & 97.9 & 76.9 & 89.8 & 96.9 & 84.1 & 84.2 & 88.6 \\
FreqNet ~\cite{tan2024frequency} & AAAI 2024 & 89.6 & 98.8 & 98.6 & 66.8 & 86.5 & 97.3 & 75.8 & 81.4 & 86.8 \\
FatFormer~\cite{liu2024forgery} & CVPR 2024 & 92.7 & \bf100.0 & \bf99.9 & 75.9 & 88.0 & \bf99.9 & \underline{98.8} & 55.8 & 88.9 \\
DRCT~\cite{chendrct} & ICML 2024 & 91.5&95.0&94.4&79.4&89.2&94.7&90.0&81.7&89.5 \\
AIDE ~\cite{yan2024sanity} & ICLR 2025 & 79.4 &  99.7 & 99.8 & 78.5 & 91.8 & 98.7 & 80.3 & 66.9 & 86.9 \\
C2P-CLIP ~\cite{tan2025c2p} & AAAI 2025 & 88.2 &  90.9 & 97.9 & \bf96.4 & \bf99.0 & 98.8 & 96.5 & \underline{98.7} & \underline{95.8} \\
SAFE $^\spadesuit$ ~\cite{li2025improving} & KDD 2025 & \bf95.7 & 99.9 & 99.8 & 59.5 & 91.7 & 95.7 & 98.0 & 77.4 & 90.3 \\
ATTSD ~\cite{cai2025adaptive}&ICCV 2025&91.8&94.7&93.8&85.0&\underline{97.3}&91.4&81.2 &90.2&90.7\\
MiraGe~\cite{shi2025mirage} & ACMMM 2025 & 83.2 & 98.8& 98.5& 82.7 &91.3& 97.6& 92.4& 96.5& 92.6 \\
\hline
\rowcolor{blue!20} Ours& -- &  \underline{95.1} & \underline{99.9} & \underline{99.8} & \underline{85.4} & 95.9 & \underline{99.8} & \bf98.8 & \bf98.7 & \textbf{96.7} \\
\bottomrule
    \end{tabular}
}
\label{tab:3}
\end{table*}

\subsection{Main Results}
We report the accuracy and average precision on the UniversalFakeDetect dataset in Tab. \ref{tab:1} and Tab. \ref{tab:2}, respectively. Our proposed method achieves SOTA results with 95.61\% mAcc and 99.32\% mAP across 19 diverse sub-test sets, outperforming all existing methods. In terms of mAcc, we surpass the SOTA methods C2P-CLIP by 1.82\%. In terms of mAP, we surpass the SOTA methods RINE by 0.53\%. Furthermore, compared with the baseline UniFD, our method brings substantial improvements of 14.23\% in mAcc and 9.18\% in mAP. These gains clearly demonstrate that our model, when trained solely on GAN-generated data \cite{goodfellow2014generative}, can generalize robustly to a wide range of challenging testing scenarios. Beyond the overall metrics, our method ranks first or second on 9 out of 19 sub-test sets in terms of mAcc. While previous approaches may excel on specific generative models but degrade considerably on others, our
method maintains consistently strong performance across almost all synthesis types, highlighting its superior generalization ability. This is mainly attributed to our adaptive design, which enables the model to capture diverse artifacts with minimal bias toward any specific generative family.

\begin{figure*}[t]
    \centering
    \includegraphics[width=0.94\textwidth]{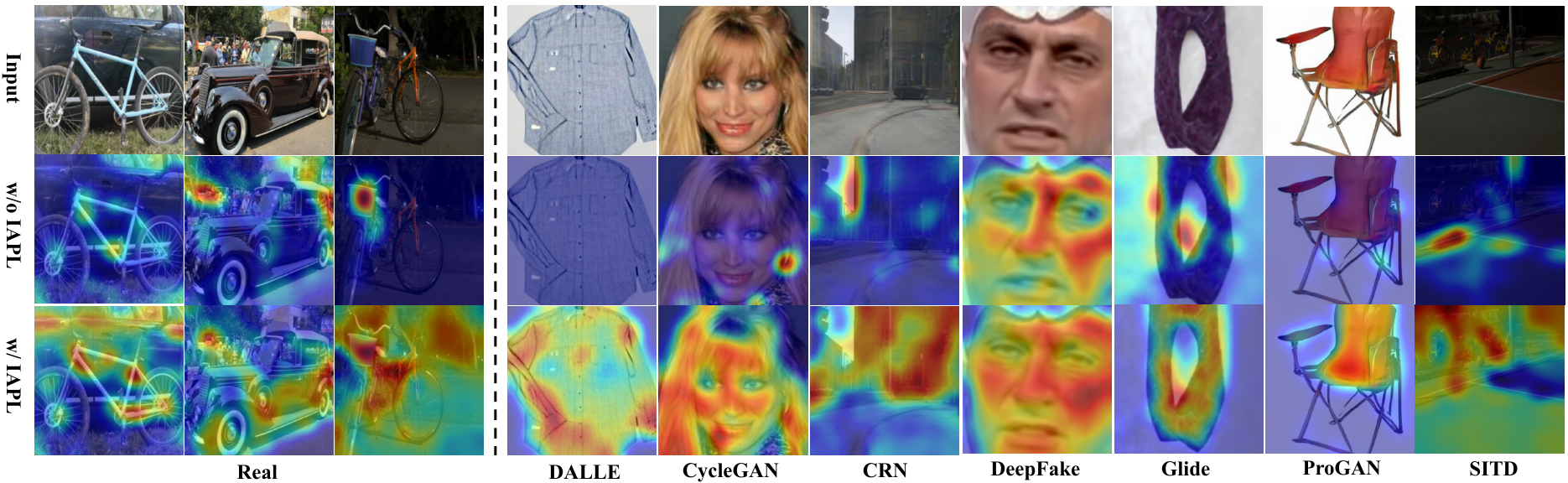}
    \caption{\textbf{Comparison of Grad-CAM beteween without (w/o) and with (w/) IAPL.} Brighter colors represent the salient region.}
    \label{fig4}
\end{figure*} 

For the GenImage dataset, we report the overall accuracy results in Tab.~\ref{tab:3}. Our method achieves 96.7\% accuracy on the full test set, significantly surpassing recent methods ATTSD and MiraGe by 6.0\% and 4.1\%, respectively. In comparison with the SOTA approach C2P-CLIP, our method further improves the mean mAcc by 0.9\%. Moreover, it outperforms the baseline UniFD by a large margin of 17.2\% in mAcc. These results demonstrate that our method, trained on a diffusion-based synthetic dataset \cite{song2020score}, exhibits strong generalization capability across diverse types of testing manipulations. Beyond the overall metrics, our method ranks first or second on 7 out of 8 sub-test sets, highlighting its stability and generalization across diverse scenarios. Results on both UniversalFake and GenImage datasets confirm the consistent performance of our method across various forgeries and generators.

\begin{table}[t]
\centering
\footnotesize
\caption{\textbf{Effectiveness of each proposed component}. RAcc and FAcc denote mAcc for real and fake images.}
\label{tab:4}
\tabcolsep=0.25cm
\resizebox{0.48\textwidth}{!}{%
\begin{tabular}{lcccc}
\toprule
Component & mAcc & mAP & RAcc & FAcc \\
\midrule
Baseline & 84.19 & 93.52 & 87.24 & 81.15 \\
+MLP-based Adapters & 89.02 & 97.66 & 86.76 & 91.28 \\
+Learnable Tokens & 91.83 & 98.04 & 96.79 & 86.87 \\
+Image-Adaptive Prompts & 94.88 & 98.72 & 97.07 & 92.68 \\
+Optimal Input Selection & 95.61 & 99.32 & 96.37 & 94.84 \\
\bottomrule
\end{tabular}
}
\end{table}

\begin{table}[t]
\centering
\footnotesize
\caption{\textbf{Ablation of Image-Adaptive Prompt Learning.} 
Default configurations are indicated with a background in \colorbox{blue!20}{blue}.}
\label{tab:5}
\tabcolsep=0.15cm
\resizebox{0.48\textwidth}{!}{
\begin{tabular}{ccccccccc}
\toprule
FSC.& GEC.& LSF.& TTT. & CFS. & mAcc & mAP & RAcc & FAcc \\
\midrule
\checkmark & & & & & 92.38&97.88&94.86&89.90\\
\checkmark & \checkmark & & & & 93.32&98.20&96.26&90.38 \\
\checkmark & \checkmark & \checkmark & & & 93.51 & 98.24 & 96.66 & 90.36 \\
\checkmark & \checkmark & \checkmark & \checkmark & & 94.58 & 98.57 & 96.94 & 92.22 \\
\rowcolor{blue!20} \checkmark & \checkmark & \checkmark & \checkmark & \checkmark & 94.88 & 98.72 & 97.07 & 92.68 \\
\bottomrule
\end{tabular}
}
\end{table}

\begin{table}[!t]
  \centering
  \footnotesize
  \begin{minipage}{0.51\linewidth}
    \centering
    \tabcolsep=0.3cm
    \captionof{table}{\textbf{Ablation of MLP-based adapter.}}
    \begin{tabular}{ccc}
      \toprule
      Config & mAcc & mAP \\
      \midrule
      $[1, 6, 1]$   & 78.27 & 95.16 \\
      $[18, 24, 1]$ & 87.74 & 96.37 \\
      \rowcolor{blue!20} $[4, 24, 4]$ & 89.02 & 97.66 \\
      $[3, 24, 3]$  & 89.37 & 97.00 \\
      \bottomrule
    \label{tab:6}
    \end{tabular}
  \end{minipage}
  \hfill
  \begin{minipage}{0.45\linewidth}
    \centering
    \tabcolsep=0.35cm
    \captionof{table}{\textbf{Number of layers $N_t$ with learnable tokens.}}
    \begin{tabular}{ccc}
      \toprule
      $N_t$ & mAcc & mAP \\
      \midrule
      12 & 89.91 & 96.96  \\
      \rowcolor{blue!20} 9 &  91.83 & 98.04  \\
      6 &  91.50 & 97.91  \\
      3 &  90.46 & 97.18 \\
      \bottomrule
    \label{tab:7}
    \end{tabular}
  \end{minipage}
  \vspace{-2mm}
\end{table}

\begin{table}[!t]
  \centering
  \footnotesize
  \begin{minipage}{0.45\linewidth}
    \centering
    \captionof{table}{\textbf{Entropy loss type.}}
    \begin{tabular}{lcc}
      \toprule
      Loss & mAcc & mAP \\
      \midrule
      \rowcolor{blue!20} Averaged & 95.61 & 99.32 \\
      Pointwise & 95.35 & 99.30 \\
      \bottomrule
    \label{tab:8}
    \end{tabular}
  \end{minipage}
  \hfill
  \begin{minipage}{0.53\linewidth}
    \centering
    \tabcolsep=0.23cm
    \captionof{table}{\textbf{Effect of tuning steps $T$.}}
    \begin{tabular}{lc>{\columncolor{blue!20}}cc}
      \toprule
      $T$ & 1 & 2 & 3 \\
      \midrule
      mAcc & 94.47 & 94.88 & 94.94 \\
      mAP  & 98.53 & 98.72 & 98.80 \\
      \bottomrule
    \label{tab:9}   
    \end{tabular}
  \end{minipage}
  \vspace{-5mm}
\end{table}

\subsection{Ablation Study}

Several ablation studies are conducted to verify the effectiveness of each key component of our method. 
All the experiments are performed on the UniversalFakeDetect dataset. We report mAcc and mAP on the testing set. 

As shown in Tab. \ref{tab:4}, we first verify the effectiveness of the four main modules of our method. It can be observed that the MLP-based adapter, learnable tokens, image-adaptive prompt, and Optimal Input Selection all significantly improve the detection performance. Using them together achieves the best performance. Importantly, even with only simple MLP-based adapters and learnable tokens, our method surpasses prior approaches with heavy architectural changes, such as Fatformer.
As shown in Tab. \ref{tab:5}, we progressively evaluate the contributions of forgery-specific condition (\textbf{FSC.}), general condition (\textbf{GEC.}), the learnable scaling factor (\textbf{LSF.}), Test-Time Token Tuning (\textbf{TTT.}), and Confidence Selection (\textbf{CFS.}). Each component consistently improves performance, with the full configuration achieving the highest accuracy, average precision, real accuracy, and fake accuracy.
As shown in Tab. \ref{tab:6}, the three numbers in the config represent the starting layer, the ending layer, and the interval between inserted adapters.
The results indicate that evenly inserting adapters into six layers yields better performance than placing them in the first six layers, the last six layers, or inserting them into more layers.
Moreover, inserting the adapter into 6 layers already yields satisfactory performance, and further increasing the number of inserted layers brings only marginal improvements. 
As shown in Tab. \ref{tab:7}, we validated the number of encoder blocks $N_t$ where learnable tokens are employed. Under the current experimental setup, the image-adaptive prompt in the first layer is replaced with learnable tokens. The results show that using 9 tokens achieves the best performance, indicating a balanced trade-off between model capacity and efficiency. Performance decreases slightly when using more or fewer learnable tokens.
As shown in Tab. \ref{tab:8}, we conduct experiments on the choice of loss used in Test-Time Token Tuning. It can be observed that using averaged entropy losses yields better results. This may be because the averaged entropy loss could be regarded as the sum of pointwise entropy loss and the KL divergence term \cite{sheng2025r}. The KL divergence term encourages consistent predictions among selected views. Improving prediction consistency across different views in forgery detection can reduce the interference of local misjudgments and enhance the model's ability to detect images with ambiguous boundaries or low levels of forgery.
As shown in Tab. \ref{tab:9}, we present the effect of varying the number of tuning steps ($T$) on model performance. As $T$ increases from 1 to 3, both accuracy Acc and AP steadily improve. Finally, we set $T=2$ for a balanced trade-off between efficiency and effectiveness.
As shown in Fig. \ref{fig4}, we visualize the Grad-CAM \cite{selvaraju2017grad} without and with IAPL. It could be observed that IAPL enables the model to focus more precisely on forgery-related regions, whereas the baseline tends to attend to irrelevant areas. 

\section{Conclusion}

In this paper, we propose Image-Adaptive Prompt Learning (IAPL), a novel strategy for generalizable AI-generated image detection. Unlike prior fine-tuning approaches that rely on fixed parameters and struggle with unseen forgeries, IAPL dynamically adapts prompts to each input image via a Conditional Information Learner and Test-Time Token Tuning. The former captures both forgery-specific and general cues, while the latter optimizes adaptive tokens during inference by enforcing consistency across augmented views, ensuring alignment with the characteristics of each image. Extensive experiments demonstrate its well performance.

\section*{Acknowledgments}
This work was supported in part by the New Generation Artificial Intelligence-National Science and Technology Major Project (No. 2025ZD0123501), Chinese National Natural Science Foundation Projects U23B2054, 62276254, 
the Science and Technology Development Fund of Macau Project 0140/2024/AGJ, the Science and Technology Foundation of Guangdong Province Project 2025A0505010025

{
    \small
    \bibliographystyle{ieeenat_fullname}
    \bibliography{main}

@String(ICLR = {Int. Conf. Learn. Represent.})

@String(AAAI = {AAAI})

@String(ICLR  = {ICLR})

@article{goodfellow2014generative,
  title={Generative adversarial nets},
  author={Goodfellow, Ian J and Pouget-Abadie, Jean and Mirza, Mehdi and Xu, Bing and Warde-Farley, David and Ozair, Sherjil and Courville, Aaron and Bengio, Yoshua},
  journal={Advances in neural information processing systems},
  volume={27},
  year={2014}
}

@article{song2020score,
  title={Score-based generative modeling through stochastic differential equations},
  author={Song, Yang and Sohl-Dickstein, Jascha and Kingma, Diederik P and Kumar, Abhishek and Ermon, Stefano and Poole, Ben},
  journal={arXiv preprint arXiv:2011.13456},
  year={2020}
}

@inproceedings{jeong2022bihpf,
  title={Bihpf: Bilateral high-pass filters for robust deepfake detection},
  author={Jeong, Yonghyun and Kim, Doyeon and Min, Seungjai and Joe, Seongho and Gwon, Youngjune and Choi, Jongwon},
  booktitle={Proceedings of the IEEE/CVF Winter Conference on Applications of Computer Vision},
  pages={48--57},
  year={2022}
}

@inproceedings{tan2024frequency,
  title={Frequency-aware deepfake detection: Improving generalizability through frequency space domain learning},
  author={Tan, Chuangchuang and Zhao, Yao and Wei, Shikui and Gu, Guanghua and Liu, Ping and Wei, Yunchao},
  booktitle={Proceedings of the AAAI Conference on Artificial Intelligence},
  volume={38},
  number={5},
  pages={5052--5060},
  year={2024}
}

@inproceedings{tan2024rethinking,
  title={Rethinking the up-sampling operations in cnn-based generative network for generalizable deepfake detection},
  author={Tan, Chuangchuang and Zhao, Yao and Wei, Shikui and Gu, Guanghua and Liu, Ping and Wei, Yunchao},
  booktitle={Proceedings of the IEEE/CVF Conference on Computer Vision and Pattern Recognition},
  pages={28130--28139},
  year={2024}
}

@inproceedings{tan2023learning,
  title={Learning on gradients: Generalized artifacts representation for gan-generated images detection},
  author={Tan, Chuangchuang and Zhao, Yao and Wei, Shikui and Gu, Guanghua and Wei, Yunchao},
  booktitle={Proceedings of the IEEE/CVF Conference on Computer Vision and Pattern Recognition},
  pages={12105--12114},
  year={2023}
}

@inproceedings{tan2025c2p,
  title={C2p-clip: Injecting category common prompt in clip to enhance generalization in deepfake detection},
  author={Tan, Chuangchuang and Tao, Renshuai and Liu, Huan and Gu, Guanghua and Wu, Baoyuan and Zhao, Yao and Wei, Yunchao},
  booktitle={Proceedings of the AAAI Conference on Artificial Intelligence},
  volume={39},
  number={7},
  pages={7184--7192},
  year={2025}
}

@inproceedings{radford2021learning,
  title={Learning transferable visual models from natural language supervision},
  author={Radford, Alec and Kim, Jong Wook and Hallacy, Chris and Ramesh, Aditya and Goh, Gabriel and Agarwal, Sandhini and Sastry, Girish and Askell, Amanda and Mishkin, Pamela and Clark, Jack and others},
  booktitle={International conference on machine learning},
  pages={8748--8763},
  year={2021},
  organization={PmLR}
}

@inproceedings{liu2024forgery,
  title={Forgery-aware adaptive transformer for generalizable synthetic image detection},
  author={Liu, Huan and Tan, Zichang and Tan, Chuangchuang and Wei, Yunchao and Wang, Jingdong and Zhao, Yao},
  booktitle={Proceedings of the IEEE/CVF Conference on Computer Vision and Pattern Recognition},
  pages={10770--10780},
  year={2024}
}

@inproceedings{lin2025standing,
  title={Standing on the shoulders of giants: Reprogramming visual-language model for general deepfake detection},
  author={Lin, Kaiqing and Lin, Yuzhen and Li, Weixiang and Yao, Taiping and Li, Bin},
  booktitle={Proceedings of the AAAI Conference on Artificial Intelligence},
  volume={39},
  number={5},
  pages={5262--5270},
  year={2025}
}

@article{zhu2023genimage,
  title={Genimage: A million-scale benchmark for detecting ai-generated image},
  author={Zhu, Mingjian and Chen, Hanting and Yan, Qiangyu and Huang, Xudong and Lin, Guanyu and Li, Wei and Tu, Zhijun and Hu, Hailin and Hu, Jie and Wang, Yunhe},
  journal={Advances in Neural Information Processing Systems},
  volume={36},
  pages={77771--77782},
  year={2023}
}

@article{yan2024sanity,
  title={A sanity check for ai-generated image detection},
  author={Yan, Shilin and Li, Ouxiang and Cai, Jiayin and Hao, Yanbin and Jiang, Xiaolong and Hu, Yao and Xie, Weidi},
  journal={arXiv preprint arXiv:2406.19435},
  year={2024}
}

@article{zhou2022learning,
  title={Learning to prompt for vision-language models},
  author={Zhou, Kaiyang and Yang, Jingkang and Loy, Chen Change and Liu, Ziwei},
  journal={International Journal of Computer Vision},
  volume={130},
  number={9},
  pages={2337--2348},
  year={2022},
  publisher={Springer}
}

@inproceedings{zhou2022conditional,
  title={Conditional prompt learning for vision-language models},
  author={Zhou, Kaiyang and Yang, Jingkang and Loy, Chen Change and Liu, Ziwei},
  booktitle={Proceedings of the IEEE/CVF conference on computer vision and pattern recognition},
  pages={16816--16825},
  year={2022}
}

@article{shu2022test,
  title={Test-time prompt tuning for zero-shot generalization in vision-language models},
  author={Shu, Manli and Nie, Weili and Huang, De-An and Yu, Zhiding and Goldstein, Tom and Anandkumar, Anima and Xiao, Chaowei},
  journal={Advances in Neural Information Processing Systems},
  volume={35},
  pages={14274--14289},
  year={2022}
}

@inproceedings{ojha2023towards,
  title={Towards universal fake image detectors that generalize across generative models},
  author={Ojha, Utkarsh and Li, Yuheng and Lee, Yong Jae},
  booktitle={Proceedings of the IEEE/CVF Conference on Computer Vision and Pattern Recognition},
  pages={24480--24489},
  year={2023}
}

@inproceedings{wang2020cnn,
  title={CNN-generated images are surprisingly easy to spot... for now},
  author={Wang, Sheng-Yu and Wang, Oliver and Zhang, Richard and Owens, Andrew and Efros, Alexei A},
  booktitle={Proceedings of the IEEE/CVF conference on computer vision and pattern recognition},
  pages={8695--8704},
  year={2020}
}

@inproceedings{chendrct,
  title={Drct: Diffusion reconstruction contrastive training towards universal detection of diffusion generated images},
  author={Chen, Baoying and Zeng, Jishen and Yang, Jianquan and Yang, Rui},
  booktitle={Forty-first International Conference on Machine Learning},
  year={2024}
}

@inproceedings{cheng2025co,
  title={CO-SPY: Combining Semantic and Pixel Features to Detect Synthetic Images by AI},
  author={Cheng, Siyuan and Lyu, Lingjuan and Wang, Zhenting and Zhang, Xiangyu and Sehwag, Vikash},
  booktitle={Proceedings of the Computer Vision and Pattern Recognition Conference},
  pages={13455--13465},
  year={2025}
}

@inproceedings{koutlis2024leveraging,
  title={Leveraging representations from intermediate encoder-blocks for synthetic image detection},
  author={Koutlis, Christos and Papadopoulos, Symeon},
  booktitle={European Conference on Computer Vision},
  pages={394--411},
  year={2024},
  organization={Springer}
}

@inproceedings{jia2022visual,
  title={Visual prompt tuning},
  author={Jia, Menglin and Tang, Luming and Chen, Bor-Chun and Cardie, Claire and Belongie, Serge and Hariharan, Bharath and Lim, Ser-Nam},
  booktitle={European conference on computer vision},
  pages={709--727},
  year={2022},
  organization={Springer}
}

@article{chen2022adaptformer,
  title={Adaptformer: Adapting vision transformers for scalable visual recognition},
  author={Chen, Shoufa and Ge, Chongjian and Tong, Zhan and Wang, Jiangliu and Song, Yibing and Wang, Jue and Luo, Ping},
  journal={Advances in Neural Information Processing Systems},
  volume={35},
  pages={16664--16678},
  year={2022}
}

@inproceedings{li2025unleashing,
  title={Unleashing the Potential of Consistency Learning for Detecting and Grounding Multi-Modal Media Manipulation},
  author={Li, Yiheng and Yang, Yang and Tan, Zichang and Liu, Huan and Chen, Weihua and Zhou, Xu and Lei, Zhen},
  booktitle={Proceedings of the Computer Vision and Pattern Recognition Conference},
  pages={9242--9252},
  year={2025}
}

@inproceedings{li2021frequency,
  title={Frequency-aware discriminative feature learning supervised by single-center loss for face forgery detection},
  author={Li, Jiaming and Xie, Hongtao and Li, Jiahong and Wang, Zhongyuan and Zhang, Yongdong},
  booktitle={Proceedings of the IEEE/CVF conference on computer vision and pattern recognition},
  pages={6458--6467},
  year={2021}
}

@inproceedings{wang2023noise,
  title={Noise based deepfake detection via multi-head relative-interaction},
  author={Wang, Tianyi and Chow, Kam Pui},
  booktitle={Proceedings of the AAAI Conference on Artificial Intelligence},
  volume={37},
  number={12},
  pages={14548--14556},
  year={2023}
}

@inproceedings{cao2022end,
  title={End-to-end reconstruction-classification learning for face forgery detection},
  author={Cao, Junyi and Ma, Chao and Yao, Taiping and Chen, Shen and Ding, Shouhong and Yang, Xiaokang},
  booktitle={Proceedings of the IEEE/CVF conference on computer vision and pattern recognition},
  pages={4113--4122},
  year={2022}
}

@inproceedings{fu2025exploring,
  title={Exploring unbiased deepfake detection via token-level shuffling and mixing},
  author={Fu, Xinghe and Yan, Zhiyuan and Yao, Taiping and Chen, Shen and Li, Xi},
  booktitle={Proceedings of the AAAI Conference on Artificial Intelligence},
  volume={39},
  number={3},
  pages={3040--3048},
  year={2025}
}

@inproceedings{haliassos2021lips,
  title={Lips don't lie: A generalisable and robust approach to face forgery detection},
  author={Haliassos, Alexandros and Vougioukas, Konstantinos and Petridis, Stavros and Pantic, Maja},
  booktitle={Proceedings of the IEEE/CVF conference on computer vision and pattern recognition},
  pages={5039--5049},
  year={2021}
}

@inproceedings{li2018ictu,
  title={In ictu oculi: Exposing ai created fake videos by detecting eye blinking},
  author={Li, Yuezun and Chang, Ming-Ching and Lyu, Siwei},
  booktitle={2018 IEEE International workshop on information forensics and security (WIFS)},
  pages={1--7},
  year={2018},
  organization={Ieee}
}

@inproceedings{shiohara2022detecting,
  title={Detecting deepfakes with self-blended images},
  author={Shiohara, Kaede and Yamasaki, Toshihiko},
  booktitle={Proceedings of the IEEE/CVF conference on computer vision and pattern recognition},
  pages={18720--18729},
  year={2022}
}

@inproceedings{chen2022self,
  title={Self-supervised learning of adversarial example: Towards good generalizations for deepfake detection},
  author={Chen, Liang and Zhang, Yong and Song, Yibing and Liu, Lingqiao and Wang, Jue},
  booktitle={Proceedings of the IEEE/CVF conference on computer vision and pattern recognition},
  pages={18710--18719},
  year={2022}
}

@inproceedings{yan2024transcending,
  title={Transcending forgery specificity with latent space augmentation for generalizable deepfake detection},
  author={Yan, Zhiyuan and Luo, Yuhao and Lyu, Siwei and Liu, Qingshan and Wu, Baoyuan},
  booktitle={Proceedings of the IEEE/CVF Conference on Computer Vision and Pattern Recognition},
  pages={8984--8994},
  year={2024}
}

@inproceedings{wang2023dire,
  title={Dire for diffusion-generated image detection},
  author={Wang, Zhendong and Bao, Jianmin and Zhou, Wengang and Wang, Weilun and Hu, Hezhen and Chen, Hong and Li, Houqiang},
  booktitle={Proceedings of the IEEE/CVF International Conference on Computer Vision},
  pages={22445--22455},
  year={2023}
}

@article{liu2021p,
  title={P-tuning v2: Prompt tuning can be comparable to fine-tuning universally across scales and tasks},
  author={Liu, Xiao and Ji, Kaixuan and Fu, Yicheng and Tam, Weng Lam and Du, Zhengxiao and Yang, Zhilin and Tang, Jie},
  journal={arXiv preprint arXiv:2110.07602},
  year={2021}
}

@article{dosovitskiy2020image,
  title={An image is worth 16x16 words: Transformers for image recognition at scale},
  author={Dosovitskiy, Alexey and Beyer, Lucas and Kolesnikov, Alexander and Weissenborn, Dirk and Zhai, Xiaohua and Unterthiner, Thomas and Dehghani, Mostafa and Minderer, Matthias and Heigold, Georg and Gelly, Sylvain and others},
  journal={arXiv preprint arXiv:2010.11929},
  year={2020}
}

@article{bahng2022exploring,
  title={Exploring visual prompts for adapting large-scale models},
  author={Bahng, Hyojin and Jahanian, Ali and Sankaranarayanan, Swami and Isola, Phillip},
  journal={arXiv preprint arXiv:2203.17274},
  year={2022}
}

@inproceedings{khattak2023maple,
  title={Maple: Multi-modal prompt learning},
  author={Khattak, Muhammad Uzair and Rasheed, Hanoona and Maaz, Muhammad and Khan, Salman and Khan, Fahad Shahbaz},
  booktitle={Proceedings of the IEEE/CVF conference on computer vision and pattern recognition},
  pages={19113--19122},
  year={2023}
}

@article{wang2020tent,
  title={Tent: Fully test-time adaptation by entropy minimization},
  author={Wang, Dequan and Shelhamer, Evan and Liu, Shaoteng and Olshausen, Bruno and Darrell, Trevor},
  journal={arXiv preprint arXiv:2006.10726},
  year={2020}
}

@inproceedings{he2015delving,
  title={Delving deep into rectifiers: Surpassing human-level performance on imagenet classification},
  author={He, Kaiming and Zhang, Xiangyu and Ren, Shaoqing and Sun, Jian},
  booktitle={Proceedings of the IEEE international conference on computer vision},
  pages={1026--1034},
  year={2015}
}

@article{hu2022lora,
  title={Lora: Low-rank adaptation of large language models.},
  author={Hu, Edward J and Shen, Yelong and Wallis, Phillip and Allen-Zhu, Zeyuan and Li, Yuanzhi and Wang, Shean and Wang, Lu and Chen, Weizhu and others},
  journal={ICLR},
  volume={1},
  number={2},
  pages={3},
  year={2022}
}

@article{fridrich2012rich,
  title={Rich models for steganalysis of digital images},
  author={Fridrich, Jessica and Kodovsky, Jan},
  journal={IEEE Transactions on information Forensics and Security},
  volume={7},
  number={3},
  pages={868--882},
  year={2012},
  publisher={IEEE}
}

@inproceedings{sheng2025r,
  title={R-TPT: Improving Adversarial Robustness of Vision-Language Models through Test-Time Prompt Tuning},
  author={Sheng, Lijun and Liang, Jian and Wang, Zilei and He, Ran},
  booktitle={Proceedings of the Computer Vision and Pattern Recognition Conference},
  pages={29958--29967},
  year={2025}
}

@article{progan,
  title={Progressive growing of gans for improved quality, stability, and variation},
  author={Karras, Tero and Aila, Timo and Laine, Samuli and Lehtinen, Jaakko},
  journal={arXiv preprint arXiv:1710.10196},
  year={2017}
}

@inproceedings{Zhu-2017-cycleGAN,
  title={Unpaired image-to-image translation using cycle-consistent adversarial networks},
  author={Zhu, Jun-Yan and Park, Taesung and Isola, Phillip and Efros, Alexei A},
  booktitle={Proceedings of the IEEE international conference on computer vision},
  pages={2223--2232},
  year={2017}
}

@article{brock2018brock,
  title={Large scale gan training for high fidelity natural image synthesis},
  author={Brock, Andrew and Donahue, Jeff and Simonyan, Karen},
  journal={arXiv preprint arXiv:1809.11096},
  year={2018}
}

@inproceedings{stylegan,
  title={A style-based generator architecture for generative adversarial networks},
  author={Karras, Tero and Laine, Samuli and Aila, Timo},
  booktitle={Proceedings of the IEEE/CVF conference on computer vision and pattern recognition},
  pages={4401--4410},
  year={2019}
}

@inproceedings{park2019SPADE,
  title={Semantic image synthesis with spatially-adaptive normalization},
  author={Park, Taesung and Liu, Ming-Yu and Wang, Ting-Chun and Zhu, Jun-Yan},
  booktitle={Proceedings of the IEEE/CVF conference on computer vision and pattern recognition},
  pages={2337--2346},
  year={2019}
}

@inproceedings{choi2018stargan,
  title={Stargan: Unified generative adversarial networks for multi-domain image-to-image translation},
  author={Choi, Yunjey and Choi, Minje and Kim, Munyoung and Ha, Jung-Woo and Kim, Sunghun and Choo, Jaegul},
  booktitle={Proceedings of the IEEE conference on computer vision and pattern recognition},
  pages={8789--8797},
  year={2018}
}

@inproceedings{roessler2019faceforensicspp,
  title={Faceforensics++: Learning to detect manipulated facial images},
  author={Rossler, Andreas and Cozzolino, Davide and Verdoliva, Luisa and Riess, Christian and Thies, Justus and Nie{\ss}ner, Matthias},
  booktitle={Proceedings of the IEEE/CVF international conference on computer vision},
  pages={1--11},
  year={2019}
}

@inproceedings{sitd,
  title={Learning to see in the dark},
  author={Chen, Chen and Chen, Qifeng and Xu, Jia and Koltun, Vladlen},
  booktitle={Proceedings of the IEEE conference on computer vision and pattern recognition},
  pages={3291--3300},
  year={2018}
}

@inproceedings{san,
  title={Second-order attention network for single image super-resolution},
  author={Dai, Tao and Cai, Jianrui and Zhang, Yongbing and Xia, Shu-Tao and Zhang, Lei},
  booktitle={Proceedings of the IEEE/CVF conference on computer vision and pattern recognition},
  pages={11065--11074},
  year={2019}
}

@inproceedings{crn,
  title={Photographic image synthesis with cascaded refinement networks},
  author={Chen, Qifeng and Koltun, Vladlen},
  booktitle={Proceedings of the IEEE international conference on computer vision},
  pages={1511--1520},
  year={2017}
}

@inproceedings{imle,
  title={Diverse image synthesis from semantic layouts via conditional imle},
  author={Li, Ke and Zhang, Tianhao and Malik, Jitendra},
  booktitle={Proceedings of the IEEE/CVF International Conference on Computer Vision},
  pages={4220--4229},
  year={2019}
}

@article{glide,
  title={Glide: Towards photorealistic image generation and editing with text-guided diffusion models},
  author={Nichol, Alex and Dhariwal, Prafulla and Ramesh, Aditya and Shyam, Pranav and Mishkin, Pamela and McGrew, Bob and Sutskever, Ilya and Chen, Mark},
  journal={arXiv preprint arXiv:2112.10741},
  year={2021}
}

@inproceedings{dalle-orig,
  title={Zero-shot text-to-image generation},
  author={Ramesh, Aditya and Pavlov, Mikhail and Goh, Gabriel and Gray, Scott and Voss, Chelsea and Radford, Alec and Chen, Mark and Sutskever, Ilya},
  booktitle={International conference on machine learning},
  pages={8821--8831},
  year={2021},
  organization={Pmlr}
}

@inproceedings{midjourney,
	title = "https://www.midjourney.com/home/",
	author = "Midjourney",
	year = "2022",
}

@inproceedings{rombach2022high,
  title={High-resolution image synthesis with latent diffusion models},
  author={Rombach, Robin and Blattmann, Andreas and Lorenz, Dominik and Esser, Patrick and Ommer, Bj{\"o}rn},
  booktitle={Proceedings of the IEEE/CVF conference on computer vision and pattern recognition},
  pages={10684--10695},
  year={2022}
}

@article{dhariwal2021diffusion,
  title={Diffusion models beat gans on image synthesis},
  author={Dhariwal, Prafulla and Nichol, Alexander},
  journal={Advances in neural information processing systems},
  volume={34},
  pages={8780--8794},
  year={2021}
}

@inproceedings{wukong,
	title = "https://xihe.mindspore.cn/modelzoo/wukong",
	author = "Wukong",
	year = "2022",
}

@inproceedings{gu2022vector,
  title={Vector quantized diffusion model for text-to-image synthesis},
  author={Gu, Shuyang and Chen, Dong and Bao, Jianmin and Wen, Fang and Zhang, Bo and Chen, Dongdong and Yuan, Lu and Guo, Baining},
  booktitle={Proceedings of the IEEE/CVF conference on computer vision and pattern recognition},
  pages={10696--10706},
  year={2022}
}

@article{kingma2014adam,
  title={Adam: A method for stochastic optimization},
  author={Kingma, Diederik P and Ba, Jimmy},
  journal={arXiv preprint arXiv:1412.6980},
  year={2014}
}

@inproceedings{selvaraju2017grad,
  title={Grad-cam: Visual explanations from deep networks via gradient-based localization},
  author={Selvaraju, Ramprasaath R and Cogswell, Michael and Das, Abhishek and Vedantam, Ramakrishna and Parikh, Devi and Batra, Dhruv},
  booktitle={Proceedings of the IEEE international conference on computer vision},
  pages={618--626},
  year={2017}
}

@inproceedings{li2025improving,
  title={Improving synthetic image detection towards generalization: An image transformation perspective},
  author={Li, Ouxiang and Cai, Jiayin and Hao, Yanbin and Jiang, Xiaolong and Hu, Yao and Feng, Fuli},
  booktitle={Proceedings of the 31st ACM SIGKDD Conference on Knowledge Discovery and Data Mining V. 1},
  pages={2405--2414},
  year={2025}
}

@inproceedings{nguyen2024laa,
  title={Laa-net: Localized artifact attention network for quality-agnostic and generalizable deepfake detection},
  author={Nguyen, Dat and Mejri, Nesryne and Singh, Inder Pal and Kuleshova, Polina and Astrid, Marcella and Kacem, Anis and Ghorbel, Enjie and Aouada, Djamila},
  booktitle={Proceedings of the IEEE/CVF Conference on Computer Vision and Pattern Recognition},
  pages={17395--17405},
  year={2024}
}

@article{yan2024orthogonal,
  title={Orthogonal Subspace Decomposition for Generalizable AI-Generated Image Detection},
  author={Yan, Zhiyuan and Wang, Jiangming and Jin, Peng and Zhang, Ke-Yue and Liu, Chengchun and Chen, Shen and Yao, Taiping and Ding, Shouhong and Wu, Baoyuan and Yuan, Li},
  journal={arXiv preprint arXiv:2411.15633},
  year={2024}
}

@inproceedings{cai2025adaptive,
  title={Adaptive Test-Time Semantic Debiasing for AI-Generated Image Detection},
  author={Cai, Yu and Tian, Jiahe and Fu, Xiaomeng and Dai, Jiao and Han, Jizhong and Lyu, Siwei},
  booktitle={Proceedings of the IEEE/CVF International Conference on Computer Vision},
  pages={1543--1552},
  year={2025}
}

@inproceedings{frank2020leveraging,
  title={Leveraging frequency analysis for deep fake image recognition},
  author={Frank, Joel and Eisenhofer, Thorsten and Sch{\"o}nherr, Lea and Fischer, Asja and Kolossa, Dorothea and Holz, Thorsten},
  booktitle={International conference on machine learning},
  pages={3247--3258},
  year={2020},
  organization={PMLR}
}

@inproceedings{shi2025mirage,
  title={MiraGe: Multimodal Discriminative Representation Learning for Generalizable AI-Generated Image Detection},
  author={Shi, Kuo and Lu, Jie and Ye, Shanshan and Zhang, Guangquan and Fang, Zhen},
  booktitle={Proceedings of the 33rd ACM International Conference on Multimedia},
  pages={353--361},
  year={2025}
}

@article{liu2026mirror,
  title={MIRROR: Manifold Ideal Reference ReconstructOR for Generalizable AI-Generated Image Detection},
  author={Liu, Ruiqi and Cui, Manni and Qin, Ziheng and Yan, Zhiyuan and Chen, Ruoxin and Han, Yi and Li, Zhiheng and Chen, Junkai and Chen, ZhiJin and Lin, Kaiqing and others},
  journal={arXiv preprint arXiv:2602.02222},
  year={2026}
}

@article{liu2025beyond,
  title={Beyond Artifacts: Real-Centric Envelope Modeling for Reliable AI-Generated Image Detection},
  author={Liu, Ruiqi and Han, Yi and Zhang, Zhengbo and Yao, Liwei and Yan, Zhiyuan and Shen, Jialiang and Chen, ZhiJin and Sun, Boyi and Weng, Lubin and Dong, Jing and others},
  journal={arXiv preprint arXiv:2512.20937},
  year={2025}
}

@article{qin2025Scaling,
  title={Scaling Up AI-Generated Image Detection with Generator-Aware Prototypes},
  author={Qin, Ziheng and Ji, Yuheng and Tao, Renshuai and Tian, Yuxuan and Liu, Yuyang and Wang, Yipu and Zheng, Xiaolong},
  journal={arXiv preprint arXiv:2512.12982},
  year={2025}
}
}


\end{document}